\documentclass[10pt,twocolumn,letterpaper]{article}

\usepackage{cvpr}      

\definecolor{cvprblue}{rgb}{0.21,0.49,0.74}
\usepackage[pagebackref,breaklinks,colorlinks,allcolors=cvprblue]{hyperref}
\usepackage{multicol}
\usepackage{multirow}
\usepackage{algpseudocode}
\usepackage{algorithm}
\usepackage{tabularx}
\usepackage{amsmath}
\usepackage{tabularray}
\makeatletter
\newcommand{\multiline}[1]{%
  \begin{tabularx}{\dimexpr\linewidth-\ALG@thistlm}[t]{@{}X@{}}
    #1
  \end{tabularx}
}
\makeatother

\def\paperID{} 
\def\confName{CVPR}
\def\confYear{2025}

\title{TrajTok: Technical Report for 2025 Waymo Open Sim Agents Challenge}

\author{Zhiyuan Zhang*, Xiaosong Jia*, Guanyu Chen, Qifeng Li, Junchi Yan$^{\dagger}$ \\
Sch. of Computer Science \& Sch. of Artificial Intelligence, Shanghai Jiao Tong University \\
* Equal Contributions \quad\quad $^{\dagger}$ 
 Correspondence Author \\
\\ 
}

\begin{document}
\maketitle

\begin{abstract}

In this technical report, we introduce TrajTok, a trajectory tokenizer for discrete next-token-prediction based behavior generation models, which combines data-driven and rule-based methods with better coverage, symmetry and robustness, along with a spatial-aware label smoothing method for cross-entropy loss. We adopt the tokenizer and loss for the SMART model and reach a superior performance with realism score of 0.7852 on the Waymo Open Sim Agents Challenge 2025. We will open-source the code in the future. 

\end{abstract}

\section{Introduction}

\begin{figure*}[h]
    \centering
    \includegraphics[width=1\linewidth]{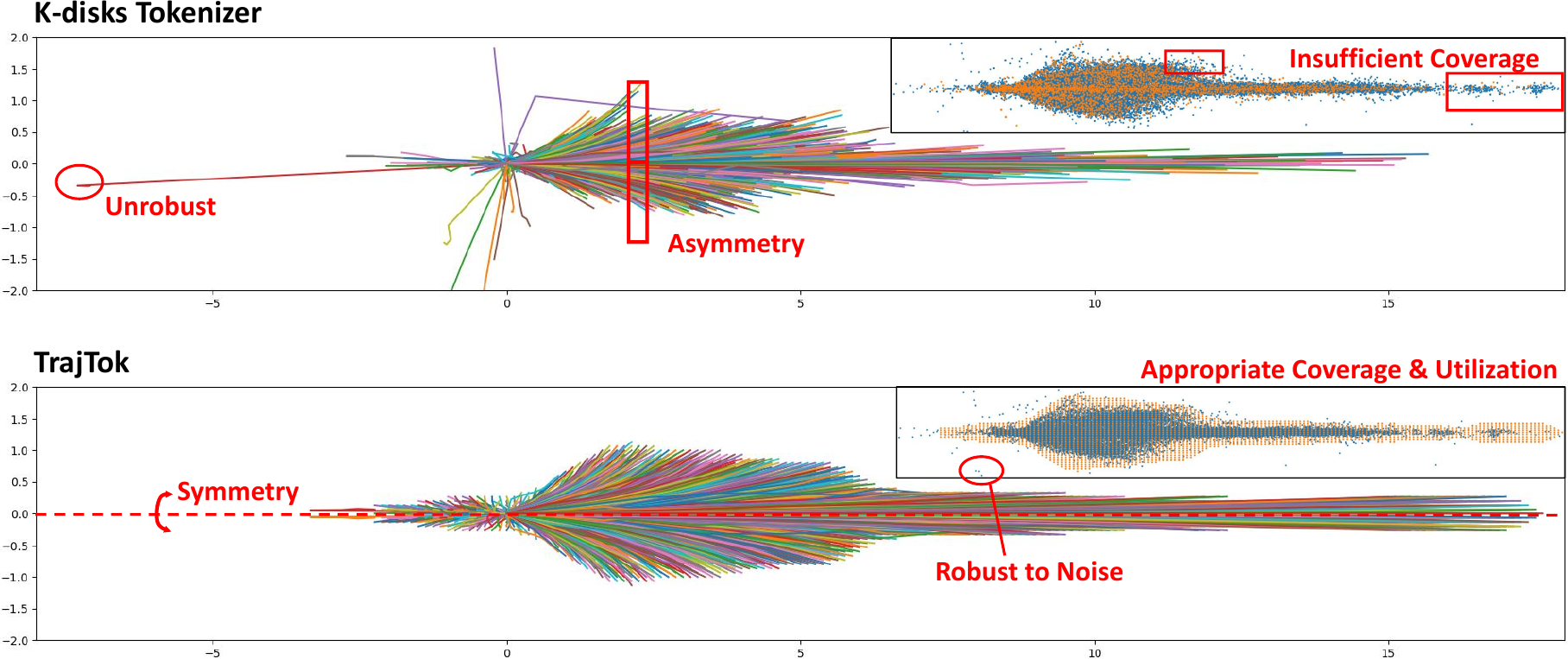}
    \caption{\textbf{Visualization and comparison of k-disks and TrajTok.} Each colored line represents a trajectory token within 0.5 seconds in agent-centric coordinate system. In the top right image, orange dots indicate the endpoints of trajectory tokens, while blue dots indicate the endpoints of real trajectories in the dataset. Compared to k-disks, TrajTok generates trajectories with better symmetry, coverage, and robustness to noise in dataset.}
    \label{fig:main}
\end{figure*}

Behavior generation is a crucial component of autonomous driving simulators. It typically takes the historical trajectories of agents and environmental information as inputs and generates future multi-agent trajectories in an autoregressive manner.

Recently, inspired by large language models, a series of behavior generation models adopt the next-token-prediction (NTP) paradigm~\citep{philion2023trajeglish, seff2023motionlm, wu2024smart, zhao2024kigras} and achieve competitive performance in the Waymo Open Sim Agents Challenge (WOSAC)~\citep{montali2023waymo}. These models use different tokenizers to generate discrete trajectory tokens from raw data in continuous space. Among them, only the k-disks tokenizer used in Trajeglish~\citep{philion2023trajeglish} and SMART~\citep{wu2024smart} is specifically designed for trajectory tokenization. However, we find that the data-driven method has poor robustness, and its coverage is limited and asymmetric, as shown in Figure~\ref{fig:main} (upper). Additionally, there is a lack of comprehensive analysis of trajectory tokenizers.

In this technical report, we propose TrajTok, a trajectory tokenizer that combines data-driven and rule-based methods, with the idea that a good trajectory tokenizer should cover more possible trajectories rather than only fit the distribution in the dataset. Our tokenizer generates trajectories with rules and filters invalid tokens with data. Specifically, it divides the trajectory space into several regions using rules, then determines valid regions based on preprocessed real trajectory data with smoothing and expanding methods, and generates one token in each region. As shown in Figure~\ref{fig:main} (lower), TrajTok offers broader coverage and better generalization and robustness compared with pure data-driven methods such as k-disks, and achieves better utilization than pure rule-based methods.

We also improve the token-related designs and propose a spatial-aware label smoothing method for cross-entropy loss for the NTP behavior generation model, further improving the performance. We replace the original k-disks tokenizer with our TrajTok on the SMART~\citep{wu2024smart} model and achieve superior performance with a realism score of 0.7852.

\section{Methods}

\subsection{Problem Formulation}
The behavior generation task can be defined as follows: Given an initial scene, including the HD map $\mathcal{M}$ and the past $T_h$ states of all agents $\{ \mathcal{S}_{-T_h}, ..., \mathcal{S}_{0} \}$, the goal is to generate the states of all agents at each future time step within ${T_f}$ steps, i.e., $\{ \mathcal{S}_{1}, ..., \mathcal{S}_{T_f} \}$. 

We focus on the discrete NTP behavior generation models that output trajectories as actions, such as SMART~\citep{wu2024smart} and Trajelish~\citep{philion2023trajeglish}. Donate the model as $\mathcal{N_{\theta,V}}$ with trainable parameters $\theta$ and a vocabulary of trajectory tokens ${\mathcal{V} = \{ c_1,c_2, ..., c_{|\mathcal{V}|}\}}$. Each trajectory token $c_i \in \mathbb{R}^{L \times 3} $ consist of $L$ points with (x, y, yaw) in the agent-centric coordinate. Donate the number of all agents as  $N_A$, the model output the trajectory tokens  $a_t \in \mathbb{R}^{N_A \times L \times 3} $ for each agent in the interval $L$ as:

\begin{equation}
    a_{t} = \mathcal{N_{\theta,V}}(\mathcal{S}_{-T_h:t \times L},\mathcal{M})
\end{equation}

where $t \in [0, 1, 2, ..., (T_f // L) ]$ is the end timestamp for each interval. The output trajectory token for each agent is select from the vocabulary, i.e. $a_{t}^{(i)} \in \mathcal{V}$. Then, a coordination transform $f$ is applied to calculate states in next interval:

\begin{equation}
\mathcal{S}_{t\times L+1:t\times (L+1)+1} = f(a_t, \mathcal{S}_t)
\end{equation}

The tokenizer $Tok()$ generates the vocabulary $\mathcal{V}$. Pure rule-based tokenizers only use predefined hyper-parameters $ \theta_{tok} $, such as the  vocabulary size and sample range as input, i.e. 
\begin{equation}
\mathcal{V}_{rule\_based} = Tok( \theta_{tok}) 
\end{equation}

Data driven tokenizers generate the vocabulary with a dataset $\mathcal{D}$, i.e.

\begin{equation}
\mathcal{V}_{data\_driven} = Tok( \theta_{tok}, \mathcal{D}) 
\end{equation}

\subsection{TrajTok}

TrajTok generates the trajectory vocabulary through the following steps:

\begin{enumerate}
    \item Data preparation and flipping. Extract all valid trajectories of length L in the dataset and normalized to the agent-centric coordinate system. Donate the normalized trajectories as $ D \in \mathbb{R}^{N_D \times L \times 3} $, where $ N_D $ is the number of trajectories. Then, flip the trajectories along the x-axis and concentrate original trajectories with flipped ones, i.e. 

    \begin{equation}
    \widetilde{D} = Concentrate( D, Flip(D))
    \label{eq:flip}
    \end{equation}    
    
    The flipping operation ensure the symmetry of trajectories, which results in symmetric tokens along the x-axis.

    \item Gridding. We generate the grid with range $x_{min}$, $x_{max}$, $y_{min}$, $y_{max}$ and interval $x_{interval}$, $y_{interval}$, which has the size of $(H, W) = (\frac{y_{max}-y_{min}}{y_{interval}},\frac{x_{max}-x_{min}}{x_{interval}})$. Each trajectory in $\widetilde{D}$ is associate with a cell if its endpoint falls within the cell. Donate $\hat{D}[i][j]$ as all trajectories associated with cell (i,j) .

    \item Filtering and expanding. A binary map $ B $ with size $(H, W)$ is generated by checking whether the trajectory points in the dataset fall within each grid cell. 
    For grid $(i, j)$ on the map, we check the value of $ B $ in a $(k \times k)$ area.
    
    The endpoints of trajectories are generally distributed in contiguous regions. If a grid cell contains trajectory points from the dataset but there are few nearby trajectories (i.e., $\hat{B}[i,j]$ is small), it indicates that the trajectory points in this cell are likely noise and should be set as invalid. Conversely, if a grid cell does not contain trajectory points from the dataset but its surrounding endpoints are dense (i.e., $\hat{B}[i,j]$ is large), then this cell is valuable for selection and should be set as valid. The filtering operation can enhance the robustness of the Tokenizer and reduce the impact of noise in the dataset, while the expansion operation can improve the coverage of tokens, enabling them to cover trajectories that do not appear in the dataset.

    \item Generating. We generate a trajectory token for each valid grid$(i, j)$ in the binary map. If there are trajectories $\hat{D}[i][j]$ belongs to the grid, the token is generated as the average trajectory:

    \begin{equation}
     c_n= Mean(\hat{D}[i][j])
    \end{equation}   

    if there are no trajectories belongs to the grid cell, which happens if the cell is set valid in expanding process, we set a endpoint $p[i][j]$  and generate a trajectory from origin point to the endpoint with curve interpolation. The x,y coordinates of the endpoint $p[i][j]$ is the center of the  grid$(i, j)$, and the yaw is estimated from the trajectories in nearby grids,i.e.
    \begin{equation}
     c_n= curve\_interp(0,p[i][j],L)
    \end{equation}

\end{enumerate}

\subsection{Spatial-Aware Label Smoothing}

NTP models often use cross-entropy loss with label smoothing to alleviate overfitting and improve generalization. Standard label smoothing assigns the same probability to each non-ground-truth label. Assuming the ground-truth label has index j, the target probability $y$ for each label can be calculated as:

\begin{equation}
y_{i}=\left\{\begin{array}{l}
1-\varepsilon \quad \text { if } \quad i=j \\
\frac{\varepsilon}{|\mathcal{V}|} \quad \text { if } \quad i \ne j
\end{array}\right.
\end{equation}

However, unlike language and image tokens, the similarity between trajectory tokens is obvious and easy to compute, and the error of the ground-truth is often within a small spatial range. Instead of equally accepting all non-ground-truth tokens, we hope the model can be more tolerant of tokens that are spatially close to the ground-truth, while rejecting tokens that are far away. Therefore, we calculate the average error between each token trajectory and the ground-truth trajectory, and assign the target probability inversely proportional to the square of the error: 
\begin{equation}
k_i = \frac{ 1}{||\mathbf{c_i}-\mathbf{c_j}||^2+\varepsilon_1}
\end{equation}
\begin{equation}
y_{i}=\left\{\begin{array}{l}
1-\varepsilon \quad \text { if } \quad i=j \\
\frac{ \varepsilon k_i}{\sum_{m=0,m \ne j}^{|\mathcal{V}|} k_m} \quad \text { if } \quad i \ne j
\end{array}\right.
\end{equation}

This spatial-aware label smoothing can both reduce errors and improve generalization.

\section{Experiments}

\subsection{Implementation Details}

\textbf{Model Settings.} We use the SMART-tiny~\citep{wu2024smart} model as base NTP behavior generation model for experiments with TrajTok. Follow original settings, the the number of decoder layers (including Map-Agent Cross-Attention Layer, Agent Interaction Self-Attention Layer and Temporal Self-Attention Layer) is 6 and the hidden dim is 128. The interval L is set to 5 and the re-plan frequency is 2 Hz, which means the model predict the 0.5-second trajectory at 10Hz every 0.5s. The original model build different trajectory vocabularies for each type of agents (vehicle, bicycle and pedestrian) but use the same head to predict classification logits for all types. We use separate prediction heads instead and set their output dim the same as the vocabulary size each. For spatial-aware label smoothing, the total target probability of all non-ground-truth labels $\varepsilon$ is 0.1, which is the same as standard label smoothing used in original model.

\textbf{Tokenizer.} For TrajTok, we set the grid range and interval for each type of agents as Table~\ref{tab:setting}. We extract trajectories that last 0.5s from the Waymo Open Motion Dataset(WOMD). The sizes of vocabularies for vehicle, bicycle and pedestrian are 8040, 3001, 2798, separately.

\begin{table}[h]
    \resizebox{\linewidth}{!}{
    \begin{tabular}{l|llllll}
    \toprule
    Agent Type  & $x_{min}$ & $x_{max}$ & $x_{interval}$ & $y_{min}$ & $y_{max}$ & $y_{interval}$ \\
    \midrule
    Vehicle  & -5 & 20 & 0.1 & -1.5 & 4.5 & 0.05 \\
    Bicycle  & -1 & 8 & 0.05 & -1 & 1 & 0.05 \\    
    Pedestrian  & -1.5 & 4.5 & 0.05 & -2 & 2 & 0.05 \\  
    \bottomrule
    \end{tabular}}
    \caption{Detailed hyper-parameters of the submit version of TrajTok. The unit of all parameters is meter.}
    \label{tab:setting}
\end{table}

\textbf{Training Details.} We train the model with 8×A100 80GB GPUs for 32 epochs on the training split of the WOMD with the AdamW optimizer. The total batch size is 48. The initial learning rate to $5 \times 10^{-4}$ and is decayed to $5 \times 10^{-6}$ based on the cosine annealing schedule.

\subsection{Main Results}

Table~\ref{tab:main_result} lists the results of top 15 entries in 2025 Waymo Open Sim Agents Challenge. TrajTok achieves 0.7852 in terms of Realism Meta metric and ranks 2nd on the leaderboard. It also reaches the state-of-the-art performance on Map-based metrics of 0.9207 and competitive performance on other metrics. Compared with baseline SMART\_topk32 which use the k-disks tokenizer, TrajTok improves the Realism Meta by +0.0038. Notably, without any fine-tuning processes, our approach has the comparable or superior performance with other methods that are also developed on the SMART-tiny model but need fine-tuning, such as SMART-tiny-CLSFT~\citep{zhang2025closed}.

\begin{table*}[t]
    \centering
    \resizebox{0.8\linewidth}{!}{
    \begin{tabular}{l|ccccc}
    \toprule
    Method  & Realism Meta $\uparrow$ & Kinematic$\uparrow$ & Interactive $\uparrow$ & Map-based $\uparrow$ & minADE $\downarrow$ \\
    \midrule
    SMART-R1 & \textbf{0.7855} & \underline{0.4940} & 0.8109 & \underline{0.9194} & 1.2990 \\
    TrajTok (Ours) & \underline{0.7852} & 0.4887 & \underline{0.8116} & \textbf{0.9207} & 1.3179 \\
    unimotion & 0.7851 & \textbf{0.4943} & 0.8105 & 0.9187 & 1.3036 \\
    SMART-tiny-CLSFT~\citep{zhang2025closed}\ & 0.7846 & 0.4931 & 0.8106 & 0.9177 & 1.3065 \\
    SMART-tiny-RLFTSim & 0.7844 & 0.4893 & \textbf{0.8128} & 0.9164 & 1.3470 \\
    comBOT & 0.7837 & 0.4899 & 0.8102 & 0.9175 & 1.3687 \\
    AgentFormer & 0.7836 & 0.4906 & 0.8103 & 0.9167 & 1.3422 \\
    UniMM~\citep{lin2025unimm} & 0.7829 & 0.4914 & 0.8089 & 0.9161 & \underline{1.2949} \\
    R1Sim & 0.7827 & 0.4894 & 0.8105 & 0.9147 & 1.3593 \\
    SimFormer & 0.7820 & 0.4920 & 0.8060 & 0.9167 & 1.3221 \\
    SMART-tiny-RLFT & 0.7815 & 0.4853 & 0.8107 & 0.9133 & 1.4266 \\
    SMART\_topk32 & 0.7814 & 0.4854 & 0.8089 & 0.9153 & 1.3931 \\
    SMART-tiny-RLFT & 0.7780 & 0.4799 & 0.8070 & 0.9109 & 1.6388 \\
    llm2ad & 0.7779 & 0.4846 & 0.8048 & 0.9109 & \textbf{1.2827} \\
    UniTFormer & 0.7776 & 0.4892 & 0.7997 & 0.9140 & 1.3592 \\

    \bottomrule
    \end{tabular}}
    \caption{Waymo Open Sim Agents Challenge leaderboard 2025. Top 15 entries in the Submission Period are presented and the Realism Meta is the primary ranking metric.}
    \label{tab:main_result}
\end{table*}

\subsection{Discussions}

\begin{table}[h]
    \resizebox{\linewidth}{!}{
    \begin{tabular}{l|lll|c}
    \toprule
    \multirow{2}{*}{Tokenizer} & \multicolumn{3}{c|}{Token Size} & \multirow{2}{*}{Realism Meta $\uparrow$} \\
    \cline{2-4}
    
    & Veh & Ped & Cyc & \\
    \midrule
    k-disks & 1024 & 1024 & 1024 & 0.7414 \\
    k-disks & 2048 & 2048 & 2048 & 0.7446 \\
    k-disks & 3072 & 3072 & 3072 & 0.7429 \\
    \midrule
    TrajTok & 2062 & 1349 & 1290 & 0.7533  \\
    TrajTok & 8040 & 3001 & 2798 & 0.7579\\
    TrajTok & 12057 & 4183 & 3923 & 0.7554 \\    

    \bottomrule
    \end{tabular}}
    \caption{The performance with different vocabulary sizes of k-disks and TrajTok. The models are trained on a subset of the training set and the Realism Meta metric uses the WOSAC 2024 version.}
    \label{tab:size}
    \vspace{-10pt}
\end{table}
\textbf{Density and vocabulary size.} Vocabulary size is an important parameter for tokenizers. Increasing the vocabulary size can improve resolution and allow for finer differentiation of trajectories, but an excessively large vocabulary may lead to model underfitting and make it more susceptible to noise. Through experiments, we found that the optimal vocabulary size for k-disks is around 2048 for each type of agent, and further increasing the vocabulary size degrades performance. For TrajTok, the optimal vocabulary size is around 8000 for vehicles and 2000-3000 for pedestrians and bicycles, as shown in Table~\ref{tab:size}.

\textbf{Coverage.} The distribution range of samples in the dataset is smaller than the range of all possible trajectories in reality. Therefore, the tokenizer should not be limited to the distribution of the dataset, so that it can cover rare cases during inference. TrajTok, through the expand method, can cover trajectories that do not appear in the dataset but may occur in reality, thus providing better generalization. Compared to naive gridding methods, it excludes a large number of trajectories that are almost impossible in the real world, improving utilization. On the other hand, the coverage of k-disks is insufficient. k-disks is more like a sampling method than a clustering method since it selects the endpoints of trajectories that appear in the dataset and generate trajectory tokens centered on them. Therefore, regardless of parameter settings, the coverage of k-disks cannot exceed that of the dataset it uses.

\textbf{Symmetry.} Considering the symmetry of vehicle kinematic models, the diversity of real-world traffic scenarios, and the left-hand and right-hand traffic rules in different regions, the candidate trajectories for an agent should be symmetric. For example, lane changes to the left and right have symmetric trajectories; in right-hand traffic regions, the turning radius for a right turn at an intersection may be smaller than that for a left turn, while in left-hand traffic regions, the opposite is the case. TrajTok ensures the symmetry of generated tokens by flipping trajectories in the dataset, whereas the tokens generated by k-disks are significantly asymmetric, which may lead to overfitting to the dataset and reduced generalization ability.

\textbf{Robustness.} The trajectories in the dataset may contain some noise and even erroneous data. Algorithms that are entirely data-driven, such as k-disks, may overfit to the noise in the data. For example, k-disks may randomly sample target points that are far behind or to the side of the current agent, which are almost impossible to reach within 0.5 seconds. In contrast, TrajTok uses a grid-based binary map filtering method, leveraging the continuity of the trajectory data distribution to effectively reduce the impact of noise in the dataset.

\section{Conclusion}

In this technical report, we propose TrajTok, a tokenizer for NTP behavior generation models, and introduce a spatial-aware label smoothing method for cross-entropy loss based on the characteristics of trajectory tokens. TrajTok focuses on a more general representation for most possible trajectories in various driving scenarios and improves coverage, symmetry, and robustness. It achieves competitive performance on the Waymo Open Sim Agents Challenge leaderboard 2025.

{
    \small
    \bibliographystyle{ieeenat_fullname}
    \bibliography{main}
}

\end{document}